%% file: arxiv.tex
\pdfoutput=1

\documentclass[11pt]{article}

\usepackage{EACL2023}

\usepackage{times}
\usepackage{latexsym}

\usepackage[T1]{fontenc}

\usepackage[utf8]{inputenc}

\usepackage{microtype}
\usepackage[utf8]{inputenc}
\usepackage{adjustbox}
\usepackage{url}
\usepackage{amssymb}
\usepackage{amsmath}
\usepackage{multirow}
\usepackage{graphicx} 
\usepackage{graphics}
\usepackage{multirow}
\usepackage{float}
\usepackage{fancyvrb}
\usepackage{amssymb}
\usepackage{amsmath}
\usepackage{amsfonts}
\usepackage{amssymb}
\usepackage{array}
\usepackage{ragged2e}
\usepackage{tabu}
\usepackage{algorithmicx}
\usepackage{algorithm}
\usepackage{algpseudocode}

\usepackage{adjustbox}
\usepackage{booktabs}

\usepackage{inconsolata}

\title{Transfer Knowledge from Natural Language to Electrocardiography:\\
Can We Detect Cardiovascular Disease Through Language Models?}

\author{Jielin Qiu$^{1}$\thanks{* marked as equal contribution}, ~William Han$^{1*}$, ~Jiacheng Zhu$^{1}$, ~Mengdi Xu$^{1}$, \\
\textbf{Michael Rosenberg$^{3}$, Emerson Liu$^{2}$, Douglas Weber$^{1}$, Ding Zhao$^{1}$  }\\
  $^{1}$Carnegie Mellon University, 
  $^{2}$Allegheny General Hospital,
  $^{3}$University of Colorado 
  }

\begin{document}
\maketitle
\begin{abstract}
Recent advancements in Large Language Models (LLMs) have drawn increasing attention since the learned embeddings pretrained on large-scale datasets have shown powerful ability in various downstream applications. 
However, whether the learned knowledge by LLMs can be transferred to clinical cardiology remains unknown. In this work, we aim to bridge this gap by transferring the knowledge of LLMs to clinical Electrocardiography (ECG). 
We propose an approach for cardiovascular disease diagnosis and automatic ECG diagnosis report generation. 
We also introduce an additional loss function by Optimal Transport (OT) to align the distribution between ECG and language embedding. 
The learned embeddings are evaluated on two downstream tasks: (1) automatic ECG diagnosis report generation, and (2) zero-shot cardiovascular disease detection. 
Our approach is able to generate high-quality cardiac diagnosis reports and also achieves competitive zero-shot classification performance even compared with supervised baselines, which proves the feasibility of transferring knowledge from LLMs to the cardiac domain.
\end{abstract}

\section{Introduction}

Heart and cardiovascular diseases are the leading global cause of death, with 80\% of cardiovascular disease-related deaths due to heart attacks and strokes. The clinical 12-lead ECG, when correctly interpreted, is the primary tool to detect cardiac abnormalities and heart-related issues. 
ECG provides unique information about the structure and electrical activity of the heart and systemic conditions through changes in the timing and morphology of the recorded waveforms. 
Achievements of ECG interpretation, such that critical and timely ECG interpretations of cardiac conditions, will lead to efficient and cost-effective intervention.

LLM starts from the Transformer model \citep{Vaswani2017AttentionIA} and grows quickly with a wide range of applications \citep{Devlin2019BERTPO,Liu2019RoBERTaAR,Brown2020LanguageMA}.
Recently, LLM has shown great potential for accelerating learning in many other domains since the learned embeddings can provide meaningful representation for downstream tasks. 
Examples include transferring the knowledge of LLM to, i.e., 
robotics control \citep{Liang2022CodeAP,Ahn2022DoAI}, 
multimodal reasoning and interaction \citep{Zeng2022SocraticMC,Zellers2021PIGLeTLG}, 
robotics planning \citep{Shah2022LMNavRN,Kant2022HousekeepTV,Jain2022TransformersAA}, 
decision-making \citep{Li2022PreTrainedLM,Huang2022LanguageMA}, robotics manipulation \citep{Shridhar2022PerceiverActorAM,Ren2022LeveragingLF,Cui2022CanFM,Tam2022SemanticEF,Khandelwal2022SimpleBE}, 
code generation \citep{Fried2022InCoderAG}, 
laws \citep{Kaplan2020ScalingLF}, 
computer vision \citep{Radford2021LearningTV}, and so on.

Some previous works explored LLM and biological protein \citep{Rives2021BiologicalSA}, or health records \citep{Yang2022GatorTronAL}.
However, the medical or healthcare domains contain so much domain knowledge that different sources preserve unique data characteristics without a unified paradigm. 
To the best of our knowledge, no previous work explores the knowledge transfer from LLM to cardiovascular disease with ECG signals.

In this work, we bridge the gap between LLM and clinical ECG by investigating the feasibility of transferring knowledge of LLM to the cardiology domain.
Our contributions are listed as follows:
\vspace{-10pt}
\begin{itemize}
    \item To the best of our knowledge, our work is the first attempt to bridge the gap between LLM and clinical cardiovascular ECG by leveraging the knowledge from pretrained LLM. 
    \vspace{-8pt}
    \item We propose a cardiovascular disease diagnosis and automatic ECG diagnosis report generation approach by transferring the knowledge from LLM to the cardiac ECG domain.
    \vspace{-8pt}
    \item We introduce an additional learning objective based on Optimal Transport distance, which empowers the model to learn the distribution between ECG and language embedding.
    \vspace{-8pt}
    \item Our method can generate high-quality cardiac diagnosis reports and achieve competitive zero-shot classification performance even compared with supervised baselines, proving the feasibility of using LLM to enhance research and applications in the cardiac domain.
\end{itemize}

\section{Related Work}

\paragraph{Cardiovascular diagnosis via ECG} The 12-lead ECG is derived from 10 electrodes placed on the surface of the skin~\citep{ecg-lead-positioning}.
An ECG works by recording electrical activity corresponding to the heartbeat muscle contractions \citep{bonow2011braunwald}.
Although computerized interpretations of ECGs are widely used, automated approaches have not yet matched the quality of expert cardiologists, leading to poor patient outcomes or even fatality~\citep{BREEN2019}.

\paragraph{Deep learning in ECG} Deep learning approaches have been rapidly adopted in many fields for their accuracy and flexibility, including ECG domain \citep{Kiranyaz2015ConvolutionalNN,pmlr-v149-nonaka21a,Khurshid2021ElectrocardiogrambasedDL,Raghunath2021DeepNN,Giudicessi2021ArtificialIA,Strodthoff2021DeepLF,AlZaiti2020MachineLP,Acharya2017ADC,Shanmugam2019MultipleIL,smigiel2021ecg}. Transformer \citep{Vaswani2017AttentionIA} has recently been adopted in several ECG applications, i.e., arrhythmia classification,  abnormalities detection, stress detection, etc  \citep{Yan2019FusingTM,Che2021ConstrainedTN,Natarajan2020AWA,Behinaein2021ATA,Song2021TransformerbasedSF,Weimann2021TransferLF}.

\paragraph{LLM in healthcare} \citet{Zhou2021NaturalLP} reviewed existing studies concerning NLP for smart healthcare. \citet{Yang2022GatorTronAL} developed a large pretrained clinical language model using transformer architecture. \citet{Steinberg2021LanguageMA} showed that using patient representation schemes inspired by techniques in LLM can increase the accuracy of clinical prediction models. More related work can be found in Appendix~\ref{sec:appendix_more_related}.


\section{Methods}

\paragraph{Problem Formulation}
We formulate the problem as generating cardiovascular diagnosis reports through pretrained LLMs. Given ECG signals $x= [x_1, x_2, ...x_t]$, our goal is to take advantage of the knowledge from LLM and learn a generated text embedding $L = [L_1, L_2, ..., L_m]$, which can then be decoded into natural language as reports or directly used for disease classification.

\begin{figure}[tp]
  \centering
  \includegraphics[width=0.99\linewidth]{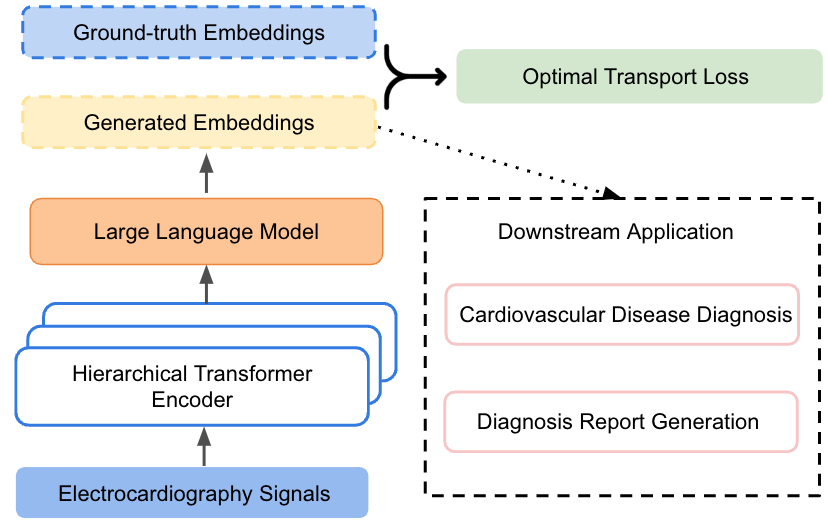}
  \caption{The architecture of our model. The Transformer encoder takes input ECG to generate ECG features as the input to LLM, where LLM transforms it into generated embeddings. An optimal transport based loss objective is formulated on generated embeddings and ground-truth embeddings for the model update.}
  \label{fig:model}
\end{figure}

\paragraph{Model Architecture} The model architecture is shown in Fig.~\ref{fig:model}, The ECG inputs are processed by hierarchical transformer encoders \citep{Vaswani2017AttentionIA} to obtain transformed ECG embeddings $X = [X_1, X_2, ..., X_n]$. Then we adopt a pretrained LLM to transform the ECG embeddings into language embeddings $L = [L_1, L_2, ..., L_m]$. For the learning objective, we use expert reports to formalize the learning loss, which includes a new loss based on Optimal Transport (OT) in addition to the traditional cross-entropy loss. The learning objective is to update the transformer encoders, which can be interpreted as a sequence-to-sequence mapping from ECG embeddings $X$ to sentence embeddings $L$. After the learning process, the learned embedding $L$ should be capable of conducting downstream applications. 

\paragraph{Downstream Applications} For the downstream applications, we first consider a classification problem that uses the embeddings $L$ for cardiovascular disease diagnosis. In addition, we consider a text generation task by decoding the output embeddings $L$ into a cardiovascular report.

\paragraph{Transformer Encoders}
The transformer is based on the attention mechanism \cite{Vaswani2017AttentionIA}. The original transformer model is composed of an encoder and a decoder. The encoder maps an input sequence into a latent representation, and the decoder uses the representation with other inputs to generate a target sequence. Our model only adopts the encoder since the target is to learn the representations of ECG features. 

The input for the Transformer is the ECG signal. First, we feed out the input into an embedding layer, which is a learned vector representation of each ECG feature by mapping each ECG feature to a vector with continuous values. Then we inject positional information into the embeddings by:
\begin{equation}\small
\begin{gathered}
P E_{(p o s, 2 i)}=\sin \left(p o s / 10000^{2 i / d_{\text {model }}}\right) \\
P E_{(p o s, 2 i+1)}=\cos \left(p o s / 10000^{2 i / d_{\text {model }}}\right)
\end{gathered}
\end{equation}
The attention model contains two sub-modules, a multi-headed attention model and a fully connected network. The multi-headed attention computes the attention weights for the input and produces an output vector with encoded information on how each feature should attend to all other features in the sequence. There are residual connections around each of the two sub-layers followed by a layer normalization, where the residual connection means adding the multi-headed attention output vector to the original positional input embedding, which helps the network train by allowing gradients to flow through the networks directly. 

In our model, our attention model contains $N$ same layers, and each layer contains two sub-layers, which are a multi-head self-attention model and a fully connected feed-forward network. Residual connection and normalization are added in each sub-layer. So the output of the sub-layer can be expressed as:
$\text{Output} = \text{LayerNorm}(x+(\text{SubLayer}(x)))$
For the Multi-head self-attention module, the attention can be expressed as:
$\text{attention} = \text{Attention}(Q,K,V)$,
where multi-head attention uses $h$ different linear transformations to project query, key, and value, which are $Q$, $K$, and $V$, respectively, and finally concatenate different attention results: 
\begin{equation}\small
\text{MultiHead(Q,K,V)} = \text{Concat}(head_1, ..., head_h) W^O
\end{equation}
\vspace{-5pt}
\begin{equation}\small
head_i = \text{Attention}(Q W^Q_i , K W^K_i , V W^V_i)
\end{equation}
where the projections are parameter matrices:
\begin{equation}\small
\begin{aligned}
&W_{i}^{Q} \in \mathbb{R}^{d_{\text {model }} d_{k}}, 
&W_{i}^{K} \in \mathbb{R}^{d_{\text {model }} d_{k}} \\
&W_{i}^{V} \in \mathbb{R}^{d_{\text {model }} d_{v}}, 
&W_{i}^{O} \in \mathbb{R}^{h d_{v} \times d_{\text {model }}}
\end{aligned}
\end{equation}
where the computation of attention adopted scaled dot-product:
\begin{equation}\small
\text{Attention}(Q,K,V) = \text{softmax} (\frac{Q K^T}{\sqrt{d_k}}) V    
\end{equation}
For the output, we use a 1D convolutional layer and softmax layer to calculate the final output.

\paragraph{Optimal Transport Loss}

OT is the problem of transporting mass between two discrete distributions supported on latent feature space $\mathcal{X}$. Let $\boldsymbol{\mu}=\left\{\boldsymbol{x}_{i}, \boldsymbol{\mu}_{i}\right\}_{i=1}^{n}$ and $\boldsymbol{v}=\left\{\boldsymbol{y}_{j}, \boldsymbol{v}_{j}\right\}_{j=1}^{m}$ be the distributions of generated embeddings and ground-truth embeddings, where $\boldsymbol{x}_{i}, \boldsymbol{y}_{j} \in \mathcal{X}$ denotes the spatial locations and $\mu_{i}, v_{j}$, respectively, denoting the non-negative masses. 
Without loss of generality, we assume $\small \sum_{i} \mu_{i}=\sum_{j} v_{j}=1$. $\pi \in \mathbb{R}_{+}^{n \times m}$ is a valid transport plan if its row and column marginals match $\mu$ and $\boldsymbol{v}$, respectively, which is $\sum_{i} \pi_{i j}=v_{j}$ and $\sum_{j} \pi_{i j}=\mu_{i}$. Intuitively, $\pi$ transports $\pi_{i j}$ units of mass at location $\boldsymbol{x}_{i}$ to new location $\boldsymbol{y}_{j}$. Such transport plans are not unique, and one often seeks a solution $\pi^{*} \in \Pi(\boldsymbol{\mu}, \boldsymbol{v})$ that is most preferable in other ways, where $\Pi(\boldsymbol{\mu}, \boldsymbol{v})$ denotes the set of all viable transport plans. OT finds a solution that is most cost-effective w.r.t. cost function $C(\boldsymbol{x}, \boldsymbol{y})$:
\begin{equation}\small
\mathcal{D}(\boldsymbol{\mu}, \boldsymbol{v})=\sum_{i j} \pi_{i j}^{*} C\left(\boldsymbol{x}_{i}, \boldsymbol{y}_{j}\right)=\inf _{\pi \in \Pi(\mu, v)} \sum_{i j} \pi_{i j} C\left(\boldsymbol{x}_{i}, \boldsymbol{y}_{j}\right)
\end{equation}
where $\mathcal{D}(\boldsymbol{\mu}, \boldsymbol{v})$ is known as OT distance. $\mathcal{D}(\boldsymbol{\mu}, \boldsymbol{v})$ minimizes the transport cost from $\boldsymbol{\mu}$ to $\boldsymbol{v}$ w.r.t. $C(\boldsymbol{x}, \boldsymbol{y})$. When $C(\boldsymbol{x}, \boldsymbol{y})$ defines a distance metric on $\mathcal{X}$, and $\mathcal{D}(\boldsymbol{\mu}, \boldsymbol{v})$ induces a distance metric on the space of probability distributions supported on $\mathcal{X}$, it becomes the Wasserstein Distance (WD). We use WD as one loss objective, in addition to the standard cross-entropy loss, for the model update. 

\section{Dataset and Prepossessing}

\paragraph{Dataset}
We conducted the experiments on the PTB-XL dataset \citep{Wagner2020PTBXLAL}, which contains clinical 12-lead ECG signals of 10-second length. There are five conditions in total, including Normal ECG (NORM), Myocardial Infarction (MI), ST/T Change (STTC), Conduction Disturbance (CD), and Hypertrophy (HYP). The waveform files are stored in WaveForm DataBase (WFDB) format with 16-bit precision at a resolution of 1$\mu$V/LSB and a sampling frequency of 100Hz. The ECG statements conform to the SCP-ECG standard and cover diagnostic, form, and rhythm statements.

\begin{figure}[H]
\centering
    \centering
	\includegraphics[width=0.48\textwidth]{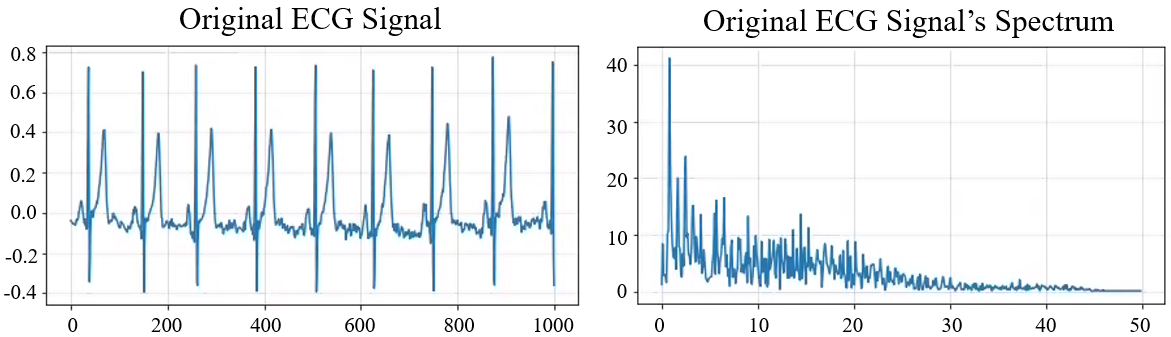}
	\caption{ECG data in the format of time series and spectrum.}
	\label{fig:fft}
\end{figure}
\begin{figure}[H]
	\centering
	\includegraphics[width=0.48\textwidth]{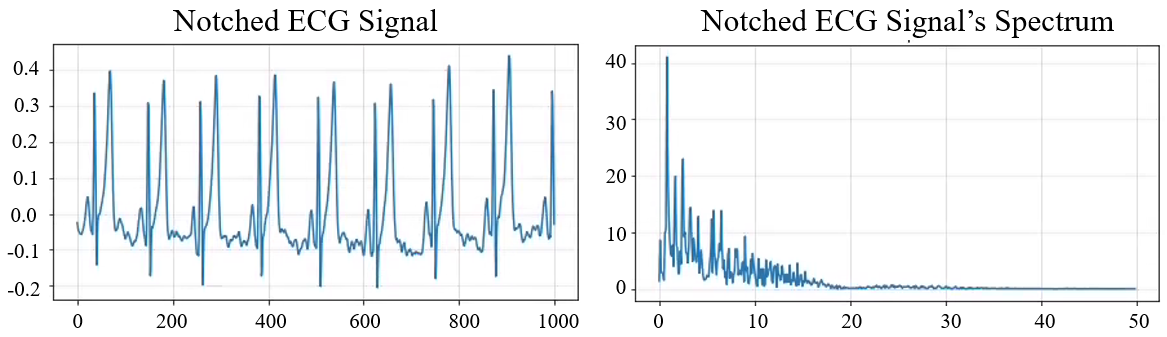}
	\caption{Filtered  ECG data  in the format of time series and spectrum.}
	\label{fig:notch}
\end{figure}
\begin{figure}[H]
\centering
	\centering
	\includegraphics[width=0.48\textwidth]{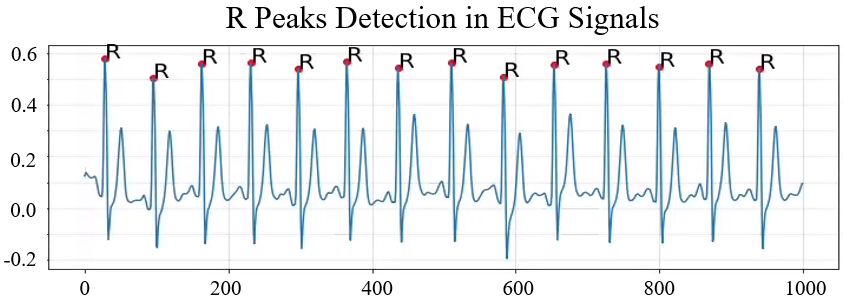}
	\caption{Detecting R peaks in the ECG signals.}
	\label{fig:find_R}
\end{figure}
\begin{figure}[H]
	\centering
	\includegraphics[width=0.48\textwidth]{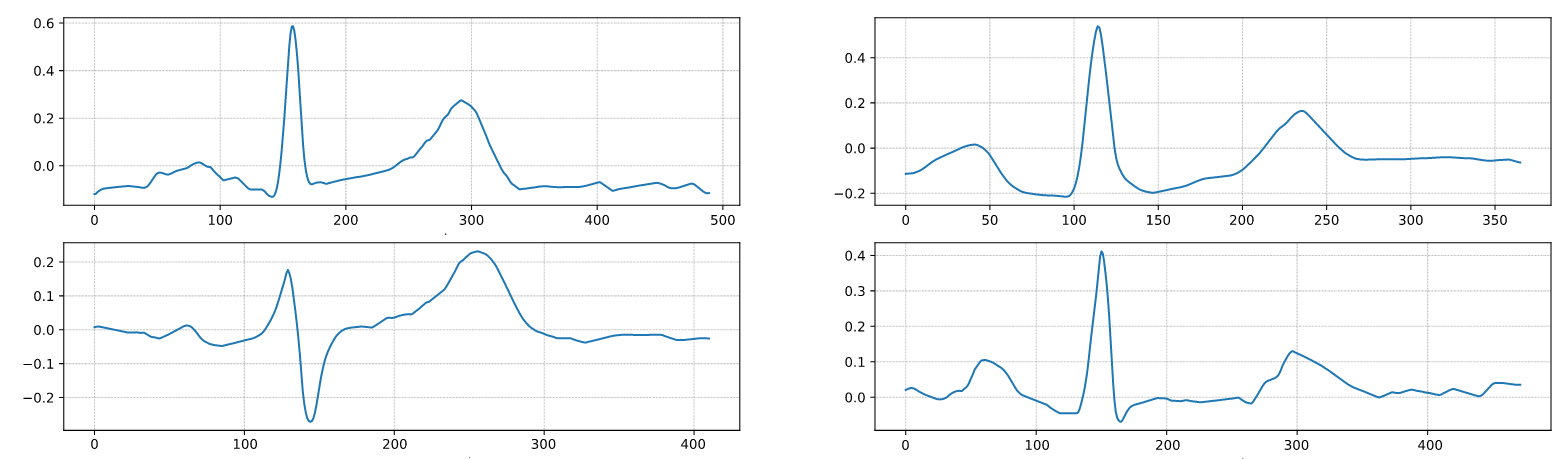}
	\caption{Extracted ECG beats divided by R peaks.}
	\label{fig:divide_R}
\end{figure}

\paragraph{Prepossessing}
The raw ECG signals are first processed by the WFDB library \citep{wfdb2022} and Fast Fourier transform (FFT) to process the time series data into the spectrum, which is shown in Fig.~\ref{fig:fft}. Then we perform n-points window filtering to filter the noise within the original ECG signals and adopt notch processing to filter power frequency interference (noise frequency: 50Hz, quality factor: 30). 
The ECG signals are segmented by dividing the 10-second ECG signals into individual ECG beats. We first detect the R peaks of each signal by ECG detectors \citep{Berndrpeak2022}, and then slice the signal at a fixed-sized interval on both sides of the R peaks to obtain individual beats. 
Examples of the filtered ECG signal results after n-points window filtering, notch processing, R peak detection, and segmented ECG beats are shown in Figures.~\ref{fig:notch},\ref{fig:find_R},\ref{fig:divide_R}.

\vspace{-20pt}
\vspace{-20pt}
\vspace{-20pt}

\paragraph{Feature Extraction}\label{sec:features} 
Instead of directly using the time-series signals,  we extract time domain and frequency domain features to better represent ECG signals. The time-domain features include: maximum, minimum, range, mean, median, mode, standard deviation, root mean square, mean square, k-order moment and skewness, kurtosis, kurtosis factor, waveform factor, pulse factor, and margin factor. The frequency-domain features include: FFT mean,  FFT variance,  FFT entropy,  FFT energy,  FFT skew,  FFT kurt,  FFT shape mean,  FFT shape std,  FFT shape skew, FFT kurt. The function of each component is shown in Table~\ref{table:fre_table}. 
An analysis of the statistics of the processed ECG data can also be found in Table~\ref{imbalance_table} in Appendix~\ref{sec:appenidx_ptb}.
\vspace{-5pt}
\begin{table}[H]\small
    \centering
	\caption{ECG statistical features in frequency domain.}
        \vspace{-5pt}
	\begin{adjustbox}{width=0.8\linewidth}
		\begin{tabular}{c|c}  
			\toprule
			Feature Symbol &Formula  \\ 
			\midrule
			$Z_1$  & $\frac{1}{N} \sum_{k=1}^{N} F(k)$  \\ 
			$Z_2$  & $\frac{1}{N-1} \sum_{k=1}^{N}\left(F(k)-Z_{1}\right)^{2}$  \\ 
			$Z_3$  & $-1 \times \sum_{k=1}^{N}\left(\frac{F(k)}{Z_{1} N} \log _{2} \frac{F(k)}{Z_{1} N}\right)$  \\ 
			$Z_4$  & $\frac{1}{N} \sum_{k=1}^{N}(F(k))^{2}$  \\ 
			$Z_5$  & $\frac{1}{N} \sum_{k=1}^{N}\left(\frac{F(k)-Z_{1}}{\sqrt{Z_{2}}}\right)^{3}$   \\ 
			$Z_6$  & $\frac{1}{N} \sum_{k=1}^{N}\left(\frac{F(k)-Z_{1}}{\sqrt{Z_{2}}}\right)^{4}$  \\ 
			$Z_7$  & $\frac{\sum_{k=1}^{N}(f(k)-F(k))}{\sum_{k=1}^{N} F(k)}$  \\ 
			 $Z_8$  & $\sqrt{\frac{\sum_{k=1}^{N}\left[\left(f(k)-Z_{6}\right)^{2} F(k)\right]}{\sum_{k=1}^{N} F(k)}}$  \\ 
			 $Z_9$ & $\frac{\sum_{k=1}^{N}\left[(f(k)-F(k))^{3} F(k)\right]}{\sum_{k=1}^{N} F(k)}$   \\ 
			 $Z_{10}$ & $\frac{\sum_{k=1}^{N}\left[(f(k)-F(k))^{4} F(k)\right]}{\sum_{k=1}^{N} F(k)}$  \\  
			 \bottomrule
	\end{tabular}
	\end{adjustbox}
	\label{table:fre_table}
\end{table}

\begin{table*}[htp]
\centering
\caption{\small{Comparisons of different backbones on Text generation (TG) and Disease detection (DD). (BERT as LLM)}}
\vspace{-8pt}
\begin{adjustbox}{width=0.88\linewidth}
\begin{tabular}{l|ccccccccc} 
\toprule
\multirow{3}{*}{Different backbones + BERT as LLM} &\multicolumn{6}{c}{Text generation (TG)} &\multicolumn{3}{c}{Disease detection (DD)}\\
\cmidrule(r){2-7} \cmidrule(r){8-10}
& \multirow{2}{*}{BLEU-1(\%)} & \multicolumn{3}{c}{ROUGE-1(\%)}
&\multirow{2}{*}{Meteor(\%)}
&\multirow{2}{*}{BertScore(\%)} & \multirow{2}{*}{Acc}
&\multirow{2}{*}{AUCROC}
&\multirow{2}{*}{F-1}  \\ 
\cmidrule(r){3-5} 
& &P  &R &F    \\
\midrule
MLP \citep{Rumelhart1986LearningIR} & 22.24 & 17.68 & 22.63 & 18.11 & 14.27 &84.68 &0.71 & 0.89 & 0.57 \\
LSTM \citep{Hochreiter1997LongSM} & 19.74 & 19.76 & 18.83 & 17.99 & 19.54 & 84.74 & 0.73 & 0.89 & 0.55 \\
ResNet \citep{He2016DeepRL} & 21.14 & 20.35 & 30.67 & 25.08 & 19.55 & 86.88 & 0.70 & 0.86 & 0.59\\
Transformer \citep{Vaswani2017AttentionIA} & \textbf{26.93} & \textbf{25.35} & \textbf{35.67} & \textbf{28.08} & \textbf{21.23} & \textbf{88.90} & \textbf{0.77} & \textbf{0.92} & \textbf{0.68} \\
\bottomrule
\end{tabular}
\end{adjustbox}
\label{table:backbone}
\end{table*}

\begin{table*}[htp]
\centering
\caption{\small{Comparisons of different LLMs on Text generation (TG) and Disease detection (DD). (Transformer as the encoder).}}
\vspace{-8pt}
\begin{adjustbox}{width=0.88\linewidth}
\begin{tabular}{l|ccccccccc} 
\toprule
\multirow{3}{*}{Different LLMs} &\multicolumn{6}{c}{Text generation (TG)} &\multicolumn{3}{c}{Disease detection (DD)}\\
\cmidrule(r){2-7} \cmidrule(r){8-10}
& \multirow{2}{*}{BLEU-1(\%)} & \multicolumn{3}{c}{ROUGE-1(\%)}
&\multirow{2}{*}{Meteor(\%)}
&\multirow{2}{*}{BertScore(\%)} & \multirow{2}{*}{Acc}
&\multirow{2}{*}{AUCROC}
&\multirow{2}{*}{F-1}  \\ 
\cmidrule(r){3-5}
& &P  &R &F   \\
\midrule
BERT \citep{Devlin2019BERTPO} & 26.93 & 25.35 & 35.67 & 28.08 & 21.23 & 88.90 & 0.77 & \textbf{0.92} & 0.68\\
BART \citep{Lewis2020BARTDS} & 27.21 & \textbf{26.12} & 35.71 & \textbf{29.56} & \textbf{24.51} & 89.61 & 0.75 & 0.88 & 0.68\\
RoBERTa \citep{Liu2019RoBERTaAR} & 27.01 & 25.31 & 36.01 & 27.88 & 22.41 & \textbf{89.72} & 0.77 & 0.89 & 0.70\\
BioClinical BERT \citep{Alsentzer2019PubliclyAC} & \textbf{27.91} & 25.41 & \textbf{36.33} & 28.42 & 23.54 & 87.21 & \textbf{0.78} & 0.89 & \textbf{0.71} \\
PubMed BERT \citep{Gu2022DomainSpecificLM} & 27.89 & 25.21 & 35.97 & 27.70 & 24.00 & 88.56 & 0.77 & 0.88 & 0.69 \\
BioDischargeSummary BERT \citep{Alsentzer2019PubliclyAC} & 26.81 & 25.32 & 35.66 & 28.10 & 21.19 & 88.90 & 0.73 & 0.85 & 0.66 \\
\bottomrule
\end{tabular}
\end{adjustbox}
\label{table:LLM}
\end{table*}

\section{Experiments}

\subsection{Experimental Settings}

\paragraph{Data and Model} The dimension of the processed ECG is 864, including 600 ECG signals and 264 time \& frequency domain features. Experiments are conducted on two NVIDIA A6000 GPUs. All the models' parameters are listed in Appendix~\ref{sec:appendix_parameters}.
\vspace{-5pt}
\paragraph{Tasks} To evaluate the learned embeddings from ECG signals, we tested the performance on two downstream applications: automatics cardiac report generation as a text generation (TG) task, and zero-shot cardiac disease detection (DD) as a multi-class classification task.
\vspace{-5pt}
\paragraph{Evaluation} For text generation evaluation, we adopted the BLEU \citep{Papineni2002BleuAM}, ROUGE \citep{Lin2004ROUGEAP}, Meteor \citep{Banerjee2005METEORAA}, and BertScore \citep{Zhang2020BERTScoreET} as evaluation metrics. We report the standard classification evaluation metrics for zero-shot cardiac disease detection: accuracy, AUCROC, and F-1 score. 


\subsection{Results}

In Table~\ref{table:backbone}, we showed the performance of both text generation and disease detection tasks with different backbone models as baselines. We found that the Transformer encoder outperforms other backbones, i.e., MLP, LSTM, and ResNet, showing Transformer encoder could be a good selection as the feature extractor.

In Table~\ref{table:supervised-DD}, we showed the performance of our zero-shot disease detection approach, compared with supervised baselines. Even though our method is in the zero-shot setting, we can already achieve the same performance with state-of-the-art supervised learning methods, demonstrating that the transferred ECG representation from LLM is already good for practical usage. We also showed some examples of generated reports compared with ground-truth reports in Table~\ref{table:appendix_report_example}.
\vspace{-8pt}
\begin{table}[H]
\centering
\caption{Comparisons with supervised baselines (DD).}
\vspace{-8pt}
\begin{adjustbox}{width=0.99\linewidth}
\begin{tabular}{l|ccc} 
\toprule
\multirow{1}{*}{Supervised learning baselines}  & \multicolumn{1}{c}{Acc}
&\multicolumn{1}{c}{AUROC}
&\multicolumn{1}{c}{F-1}   \\ 
\midrule
Transformer \citep{Zhu2022GeoECGDA} & 0.75 &0.843 &0.575 \\
CNN \citep{migiel2021ECGSC} &0.72 &0.877 &0.611 \\
SincNet \citep{Ravanelli2018SpeakerRF} &0.73 &0.84 &0.6 \\
Contrastive Learning \citep{Lan2022IntraInterSS} & -- & 0.722 & -- \\
CNN + Entropy \citep{migiel2021ECGSC} &0.76 &0.910 &0.68 \\
\midrule
Ours$_{BERT}$ & \textbf{0.77} & \textbf{0.92} & \textbf{0.68} \\
\bottomrule
\end{tabular}
\end{adjustbox}
\label{table:supervised-DD}
\end{table}
\vspace{-15pt}
\begin{table}[H]
\centering
\caption{Examples of comparison on generated reports (marked as Predicted-X) and ground-truth reports (marked as GT-X). }
\vspace{-5pt}
\begin{adjustbox}{width=0.99\linewidth}
\begin{tabular}{l|p{7cm}} 
\toprule
\multirow{1}{*}{Backbone}  & \multicolumn{1}{c}{Reports}   \\ 
\midrule
GT-1 & ``sinus rhythm left type peripheral low voltage’’\\
Predicted-1 & ``ventricular arrhythmia flatfar arrhythmia’’\\
\midrule
GT-2 & ``sinus rhythm incomplete right block otherwise normal ekg’’ \\
Predicted-2 & ``ventricularear extrasystole block sinus rhythm or normal.’’	\\
\bottomrule
\end{tabular}
\end{adjustbox}
\label{table:appendix_report_example}
\end{table}

\subsection{Ablation Study}

To further analyze the components, we conduct ablation studies on different LLMs and the number of transformer layers (with BERT as LLM). 
Table~\ref{table:LLM} shows the results of different LLMs for the text generation and disease detection tasks. 
We found that all LLMs showed good performance in both tasks, demonstrating that knowledge can be transferred from the language domain to the cardiac domain without constraints. BART shows good performance in the text generation task, while BioClinical BERT shows better performance in the disease detection task, though the variation between different LLMs is not very large. In addition, as in Table~\ref{table:layers}, we found that more layers could lead to better representations, achieving better performance for downstream applications.
\vspace{-5pt}
\begin{table}[H]
\centering
\caption{Ablation study of different transformer layers.}
\vspace{-8pt}
\begin{adjustbox}{width=0.99\linewidth}
\begin{tabular}{l|ccccccccc} 
\toprule
\multirow{3}{*}{Layers} &\multicolumn{6}{c}{Text generation (TG)} &\multicolumn{3}{c}{Disease detection (DD)}\\
\cmidrule(r){2-7} \cmidrule(r){8-10}
& \multirow{2}{*}{BLEU-1(\%)} & \multicolumn{3}{c}{ROUGE-1(\%)}
&\multirow{2}{*}{Meteor(\%)}
&\multirow{2}{*}{BertScore(\%)} & \multirow{2}{*}{Acc}
&\multirow{2}{*}{AUCROC}
&\multirow{2}{*}{F-1}  \\ 
\cmidrule(r){3-5}
& &P  &R &F    \\
\midrule
1 & 25.81 & 20.36 & 30.72 & 23.12 & 21.38 & 83.58 & 0.69 & 0.83 & 0.59 \\
2 & 24.77 & 19.22 & 28.55 & 24.51 & 20.44 & 82.89 & 0.72 & 0.81 & 0.61 \\
3 & 25.44 & 20.44 & 27.21 & 24.81 & 19.99 & 84.63 & 0.75 & 0.80 & 0.62 \\
4 & 25.12 & 21.36 & 30.88 & 25.76 & \textbf{22.68} & 86.35 & 0.74 & 0.80 & 0.64 \\
5 & \textbf{26.93} & \textbf{25.35} & \textbf{35.67} & \textbf{28.08} & 21.23 & \textbf{88.90} & \textbf{0.77} & \textbf{0.92} & \textbf{0.68} \\
\bottomrule
\end{tabular}
\end{adjustbox}
\label{table:layers}
\end{table}

\section{Conclusion}
In this paper, we bridge the gap between LLMs and cardiovascular ECG by transferring knowledge of LLMs into the cardiovascular domain. The transferred knowledge embeddings can be used for downstream applications, including cardiovascular disease diagnosis and automatic ECG diagnosis report generation. 
Our results demonstrate the effectiveness of knowledge transfer, as the proposed method shows excellent performance in both downstream tasks, where our zero-shot classification approach even achieved competitive performance with supervised learning baselines, showing the feasibility of using LLM to enhance applications in the cardiovascular domain.

\clearpage

\section{Limitations}

Due to the constrain of the available datasets, we only conducted experiments on the PTB-XL dataset, which is the current largest ECG dataset that contains high-quality clinical ECG signals and cardiac reports by experienced cardiologists. 

We understand that collecting high-quality clinical data is much more complicated and time-consuming than collecting other data from online resources, like images, since it requires expert domain knowledge and is limited by many privacy regulations. We are working with cardiologists, hospitals, and clinical research labs, hope we can release a new dataset to provide additional materials for this research direction.

\section{Ethics Statement}

In this work, the data used as experimental materials are from publicly available databases, where the patients' information is anonymized. 
To the best of our knowledge, we do not foresee any harmful uses of this study.

\section{Acknowledgements}

The research is partially supported by the DARPA ADAPTER program, and partially supported by the Allegheny Health Network and Mario Lemieux Center
for Innovation and Research in EP.

\bibliography{anthology,custom}
\bibliographystyle{acl_natbib}

\clearpage
\appendix
\input{appendix}

\end{document}

%% file: appendix.tex
\section{More Analysis of the Dataset}\label{sec:appenidx_ptb}

\begin{table}[htp]\small
    \centering
	\caption{Statistics of the processed ECG data.}
	\vspace{-5pt}
	\begin{adjustbox}{width=0.98\linewidth}
		\begin{tabular}{c|c|c|c|c}
			\toprule
			Category & Patients & Percentage & Beats &Percentage \\   \midrule 
			NORM  &9528  &34.2\%  & 28419 &36.6\% \\ 
			MI  &5486  &19.7\%  &10959 &14.1\% \\ 
			STTC  &5250  &18.9\%   & 8906 &11.5\%   \\ 
			CD  &4907  &17.6\%  & 20955  &27.0\%  \\ 
			HYP  &2655  &9.5\%  & 8342  &10.8\%  \\ 
			\bottomrule
	\end{tabular}
	\end{adjustbox}
	\label{imbalance_table}
\end{table}

\section{Experiment Parameters}\label{sec:appendix_parameters}
We provide the experimental parameters of the models in the paper in Table~\ref{Table:exp_param} and Table~\ref{Table:base_param}.

\begin{table*}[htp]\small
\centering
\caption{Experiment parameters (best ones marked in bold).}
\vspace{-5pt}
\begin{adjustbox}{width=0.99\linewidth}
\begin{tabular}{lcccccr}
\toprule
Task & Batch Size&Encoder Layers&Att. Heads &Dropout&Epochs&Warmup Steps  \\ 
\midrule
Text Generation
& [8, \textbf{16}, 32, 64]
& [1, 2, 3, 4, \textbf{5}]
& [1, 2, 3, 4, \textbf{5}]
&[0.1, 0.2, \textbf{0.3}]
&[10, 20, \textbf{50}, 100, 200]
&[1000, \textbf{2000}]
 \\
Disease Detection
& [8, \textbf{16}, 32, 64]
& [1, 2, 3, 4, \textbf{5}]
& [1, 2, 3, 4, \textbf{5}]
&[0.1, 0.2, \textbf{0.3}]
&[10, 20, \textbf{50}, 100, 200]
&[1000, \textbf{2000}]
 \\
\bottomrule
\end{tabular}
\end{adjustbox}
\label{Table:exp_param}
\end{table*}

\begin{table*}[htp]\small
\centering
\caption{Baseline parameters (best ones marked in bold).}
\vspace{-5pt}
\begin{adjustbox}{width=0.99\linewidth}
\begin{tabular}{lcccccccr}
\toprule
Models & Batch Size&Layers&In Channel Size&Kernel Sizes&Dropout&Epochs&Warmup Steps  \\ 
\midrule
MLP
& [8, \textbf{16}, 32, 64]
& [\textbf{2}, 3, 4]
&[\textbf{128}, 256, 512, 1024]
& [\textbf{1},3]
&[0.1, 0.2, \textbf{0.3}]
&[10, 20, \textbf{50}, 100, 200]
&[1000, \textbf{2000}]
 \\
LSTM
& [8, \textbf{16}, 32, 64]
& [1, \textbf{2}, 3, 4]
&[128, \textbf{256}, 512, 1024]
& [\textbf{1},3]
&[0.1, 0.2, \textbf{0.3}]
&[10, 20, \textbf{50}, 100, 200]
&[1000, \textbf{2000}]
 \\
Resnet
& [8, \textbf{16}, 32, 64]
& [1, \textbf{2}, 3, 4]
&[128, \textbf{256}, 512, 1024]
& [\textbf{1},3]
&[0.1, 0.2, \textbf{0.3}]
&[10, 20, \textbf{50}, 100, 200]
&[1000, \textbf{2000}]
 \\
Transformer
& [8, \textbf{16}, 32, 64]
& [1, 2, 3, 4, \textbf{5}]
&[128, \textbf{256}, 512, 1024]
& [\textbf{1},3]
&[0.1, 0.2, \textbf{0.3}]
&[10, 20, \textbf{50}, 100, 200]
&[1000, \textbf{2000}]
 \\
\bottomrule
\end{tabular}
\end{adjustbox}
\label{Table:base_param}
\end{table*}

\section{More Related Work}\label{sec:appendix_more_related}

\paragraph{Cardiovascular Disease in Current Practice} Patients presenting with chest pain to the emergency department (ED) constitute a diagnostic and logistic challenge as chest pain can be caused by an extensive variety of disorders \cite{Amsterdam2010TestingOL}. Diagnostic tests and decision algorithms play a critical role in speeding up the appropriate triage of chest pain patients in the ED, facilitating further (often more invasive) testing if warranted, and preventing unnecessary hospitalization of patients with non-critical disorders. In current practice, about half of the patients presenting with chest pain can be discharged from the ED, and only 5.5 percent of all ED visits lead to serious diagnoses \cite{Hsia2016ANS}. However, research suggests the diagnosis of chest pain in the ED now costs an estimated \$10 to \$12 billion per year in the U.S. So a automatic cardiovascular disease diagnosis system is essential to provide cost-efficient patient care.

\paragraph{Deep learning in ECG} Deep learning approaches have been rapidly adopted across a wide range of fields due to their accuracy and flexibility but require large labeled training sets.
With the development in machine learning, many models have been applied to ECG disease detection \citep{Kiranyaz2015ConvolutionalNN,pmlr-v149-nonaka21a,Khurshid2021ElectrocardiogrambasedDL,Raghunath2021DeepNN,Giudicessi2021ArtificialIA,Strodthoff2021DeepLF,Qiu2022OptimalTB,Zhu2022GeoECGDA}. \citet{AlZaiti2020MachineLP} predicted acute myocardial ischemia in patients with chest pain with a fusion voting method. \citet{Acharya2017ADC,Moody2001TheIO} proposed a nine-layer deep convolutional neural network (CNN) to classify heartbeats in the MIT-BIH Arrhythmia database. \citet{Shanmugam2019MultipleIL} estimate a patient’s risk of cardiovascular death after an acute coronary syndrome by a multiple instance learning framework. Recently, \citet{smigiel2021ecg} proposed models based on SincNet \citep{Ravanelli2018SpeakerRF} and used entropy-based features for cardiovascular diseases classification. The transformer model has also recently been adopted in several ECG applications, i.e., arrhythmia classification,  abnormalities detection, stress detection, etc  \citep{Yan2019FusingTM,Che2021ConstrainedTN,Natarajan2020AWA,Behinaein2021ATA,Song2021TransformerbasedSF,Weimann2021TransferLF}. 

\paragraph{Multimodal Learning}

Formalized multimodal learning research dates back to 1989, when \citet{yuhas_1989_integration} conducted an experiment that built off the McGurk Effect for audio-visual speech recognition using neural networks \citep{tiippana_2014_what,McGurk1976HearingLA}. Aligning representations from different modalities is an important step in multimodal learning. 
With the recent advancement in computer vision and natural language processing, multimodal learning, which aims to explore the explicit relationship between vision and language, has drawn significant attention \citep{Wang2020AnEF}. There are many methods proposed for exploring the multimodal alignment objective. \citet{Torabi2016LearningLE,Yu2017EndtoEndCW} adopted attention mechanisms, \citet{Dong2021DualEF,Qiu2022LiveSegUM,Qiu2022SemanticsConsistentCS,Qiu2022MHMSMH} composed pairwise joint representation, \citet{Chen2020FineGrainedVR,Wray2019FineGrainedAR,Zhang2018CrossModalAH} learned fine-grained or hierarchical alignment, \citet{Lee2018StackedCA,Wu2019UnifiedVE} decomposed the images and texts into sub-tokens, \citet{Velickovic2018GraphAN,Yao2018ExploringVR} adopted graph attention for reasoning, and \citet{Yang2021TACoTC} applied contrastive learning algorithms for video-text alignment. 

\paragraph{Multimodal Learning in Healthcare Applications}

Many previous works have explored multimodal learning to boost performance in clinical healthcare applications, i.e., affective computing for depression disease detection and so on
\citep{Liu2021ComparingRP,Qiu2018MultiviewER,Liu2019MultimodalER,Qiu2018DataEV,Qiu2018EmotionRB,Qiu2019VisualSL,Han2022AnEE}. \citet{Liu2021ComparingRP,Qiu2018MultiviewER,Liu2019MultimodalER,Qiu2018DataEV,Qiu2018EmotionRB} explored the inner correlation between different modalities. \citet{Bao2019InvestigatingSD} investigated the demographics, showing that the subject's individual characteristics can also be involved in robustness and personalized design.  \citet{Qiu2019VisualSL} investigated the relationship between computational vision models and computational neuroscience. \citet{Hollenstein2021DecodingEB,Han2022AnEE} explored the connectivity between natural language and EEG signals.

\section{More Experiment Results}

For the results in the paper, we used ECG signals along with ECG time \& frequency domain features as inputs. To compare the performance, we also conducted the experiments by only using  ECG signals as inputs, with no time \& frequency domain features. This set of experiments can be considered an additional ablation study for the inputs. The results are shown in Tables~\ref{table:appendix_backbone_tg},\ref{table:appendix_backbone_dd},\ref{table:appendix_trans_layers_tg},\ref{table:appendix_trans_layers_dd}. 

Compare Table~\ref{table:appendix_backbone_tg} \&~\ref{table:appendix_backbone_dd} with Table~\ref{table:backbone} in the paper, we can find that the performance of only using ECG signals as inputs is lower than combining time \& frequency features as inputs in both text generation and disease detection tasks, which demonstrates that incorporating  time \& frequency features is useful for capturing the characteristics of ECG and can lead to better representations through LLM. 

In Table~\ref{table:appendix_trans_layers_tg},\ref{table:appendix_trans_layers_dd}, the transformer backbone  performs the best compared to others in both disease detection and text generation tasks, which is in consistent with the findings in the paper, showing that more layers could lead to better representations, achieving better performance for downstream applications. In addition, compared with Table~\ref{table:layers} in the paper, we can find that the performance in Table~\ref{table:appendix_trans_layers_tg} and \ref{table:appendix_trans_layers_dd} are lower than the ones in Table~\ref{table:layers}, which also proved the same findings that adding time \& frequency features is useful for learning the cardiac ECGs.

\begin{table}[htp]
\centering
\caption{Comparisons with different backbones on the text generation task, where BERT is used as LLM.}
\vspace{-5pt}
\begin{adjustbox}{width=0.99\linewidth}
\begin{tabular}{l|cccccc} 
\toprule
\multirow{2}{*}{Backbone} & \multicolumn{1}{c}{BLEU-1(\%)} & \multicolumn{3}{c}{ROUGE-1(\%)}
&\multicolumn{1}{c}{Meteor(\%)}
&\multicolumn{1}{c}{BertScore}   \\ 
\cmidrule(r){3-5}
& &P  &R &F & &F   \\
\midrule
MLP & 18.16 & 16.19 & 13.71 & 14.48 & 12.11 &80.77 \\
LSTM & 19.72 & 19.67&18.83 &17.99 &19.54 &84.73\\
Resnet & 21.15 & 20.35 & 20.67 & 24.08 & 19.55 & 85.22\\
Transformer & \textbf{24.51} & \textbf{23.22} & \textbf{30.81} & \textbf{26.19} & \textbf{20.02} & \textbf{85.44}\\
\bottomrule
\end{tabular}
\end{adjustbox}
\label{table:appendix_backbone_tg}
\end{table}

\begin{table}[H]
\centering
\caption{Comparisons with different backbones on the disease detection task, where BERT is used as LLM.}
\vspace{-5pt}
\begin{adjustbox}{width=0.6\linewidth}
\begin{tabular}{l|ccc} 
\toprule
\multirow{1}{*}{Backbone}  & \multicolumn{1}{c}{Acc}
&\multicolumn{1}{c}{AUCROC}
&\multicolumn{1}{c}{F-1}   \\ 
\midrule
MLP  &0.69 & 0.77 & 0.49\\
LSTM & 0.71 & 0.82 & 0.59\\
Resnet & 0.70 & \textbf{0.83} & 0.55\\
Transformer & \textbf{0.75} & 0.81 & \textbf{0.60}\\
\bottomrule
\end{tabular}
\end{adjustbox}
\label{table:appendix_backbone_dd}
\end{table}

\begin{table}[H]
\centering
\caption{Comparisons of different number of transformer layers on the text generation task, where BERT is used as LLM.}
\vspace{-5pt}
\begin{adjustbox}{width=0.99\linewidth}
\begin{tabular}{l|cccccc} 
\toprule
\multirow{2}{*}{LLM} & \multicolumn{1}{c}{BLEU-1(\%)} & \multicolumn{3}{c}{ROUGE-1(\%)}
&\multicolumn{1}{c}{Meteor(\%)}
&\multicolumn{1}{c}{BertScore(\%)}   \\ 
\cmidrule(r){3-5}
& &P  &R &F & &F   \\
\midrule
1 & 25.52 & 19.10 & 27.65 & 21.43 & 20.11 & 86.52 \\
2 & 24.21 & 20.00 & 28.75 & 23.90 & 20.32 & 84.66\\
3 & 23.44 & 20.44 & 27.21 & 24.81 & 19.99 & 84.63\\
4 & 23.17 & 20.99 & 28.01 & 24.44 & 20.18 & 87.65\\
5 & \textbf{25.69} & \textbf{24.75} & \textbf{34.81} & \textbf{27.59} & \textbf{21.03} & \textbf{87.33}\\
\bottomrule
\end{tabular}
\end{adjustbox}
\label{table:appendix_trans_layers_tg}
\end{table}

\begin{table}[H]
\centering
\caption{Comparisons of different number of transformer layers on the disease detection task, where BERT is used as LLM..}
\vspace{-5pt}
\begin{adjustbox}{width=0.6\linewidth}
\begin{tabular}{l|ccc} 
\toprule
\multirow{1}{*}{Num of Layers}  & \multicolumn{1}{c}{Acc}
&\multicolumn{1}{c}{AUCROC}
&\multicolumn{1}{c}{F-1}   \\ 
\midrule
1 & 0.62 & 0.79 & 0.51 \\
2 & 0.74 & 0.80 & 0.60\\
3 & 0.71 & 0.82 & 0.59\\
4  & 0.72 & 0.83 & 0.61\\
5 & \textbf{0.75} & \textbf{0.88} & \textbf{0.64}\\
\bottomrule
\end{tabular}
\end{adjustbox}
\label{table:appendix_trans_layers_dd}
\end{table}

